\def\year{2022}\relax
\documentclass[letterpaper]{article} 
\usepackage{aaai22}  
\usepackage{times}  
\usepackage{helvet}  
\usepackage{courier}  
\usepackage[hyphens]{url}  
\usepackage{graphicx} 
\usepackage{natbib}  
\usepackage{caption} 
\DeclareCaptionStyle{ruled}{labelfont=normalfont,labelsep=colon,strut=off} 
\usepackage{algorithm}
\usepackage{algorithmic}

\usepackage{amsmath}
\usepackage{subfigure}
\usepackage{makecell}
\usepackage{multirow}
\usepackage{amssymb}
\usepackage{newfloat}
\usepackage{listings}

\floatstyle{ruled}
\newfloat{listing}{tb}{lst}{}
\floatname{listing}{Listing}
\title{MDS-Net: A Multi-scale Depth Stratification Based Monocular 3D Object Detection Algorithm}
\author {
    Zhouzhen Xie\textsuperscript{\rm 1},
    Yuying Song\textsuperscript{\rm 1},
    Jingxuan Wu\textsuperscript{\rm 1},
    Zecheng Li\textsuperscript{\rm 1},
    Chunyi Song\textsuperscript{*,\rm 1,\rm 2},
    Zhiwei Xu\textsuperscript{\rm 1,\rm 2}
}
\affiliations {
    \textsuperscript{\rm 1} Zhejiang University, China\\
    \textsuperscript{\rm 2} Donghai Laboratory, China\\
\{zhouzhenxie, yysong, jxwu, lizechng, cysong, xuzw\}@zju.edu.cn
}

\usepackage{bibentry}

\begin{document}

\maketitle

\begin{abstract}
   Monocular 3D object detection is very challenging in autonomous driving due to the lack of depth information. This paper proposes a one-stage monocular 3D object detection algorithm based on multi-scale depth stratification, which uses the anchor-free method to detect 3D objects in a per-pixel prediction. In the proposed MDS-Net, a novel depth-based stratification structure is developed to improve the network's ability of depth prediction by establishing mathematical models between depth and image size of objects. A new angle loss function is then developed to further improve the accuracy of the angle prediction and increase the convergence speed of training. An optimized soft-NMS is finally applied in the post-processing stage to adjust the confidence of candidate boxes. Experiments on the KITTI benchmark show that the MDS-Net outperforms the existing monocular 3D detection methods in 3D detection and BEV detection tasks while fulfilling real-time requirements.
\end{abstract}

\section{Introduction}
3D object detection is a fundamental function to enable more complex and advanced autonomous driving tasks, such as object tracking~\cite{hu2019joint,gonzalez2020smat,weng20203d,Junhao_Zhang} and event detection~\cite{nguyen2018graph,ravanbakhsh2017abnormal}. Nowadays, most 3D object detection algorithms use the LiDAR point cloud~\cite{VoxelNet:_End-to-End_Learning_for_Point_Cloud_Based_3D_Object_Detection,Frustum_PointNets_for_3D_Object_Detection_From_RGB-D_Data,PointNet:_Deep_Learning_on_Point_Sets_for_3D_Classification_and_Segmentation,sassd,pointgnn} to provide distance information. However, the laser is usually expensive, and hard to apply in real-time for it generates large amounts of data. Furthermore, the irregularity of point clouds increases the difficulty of using convolutional operations to extract features. By contrast, the camera is cheaper and feature extraction from the image is easier to realize using the convolutional neural network. However, due to the lack of reliable depth information, 3D object detection based on monocular images is generally considered more challenging.

The critical challenge in monocular 3D object detection lies in accurately predicting object’s depth and angle in the 3D coordinate system. The existing monocular 3D object detection algorithms either generate pseudo point clouds~\cite{Monocular_3D_Object_Detection_with_Pseudo-LiDAR_Point_Cloud,Multi-Level_Fusion_Based_3D_Object_Detection_From_Monocular_Images,Accurate_Monocular_3D_Object_Detection_via_Color-Embedded_3D_Reconstruction_for_Autonomous_Driving} or directly process images to realize the detection of 3D objects~\cite{Monocular_3D_Object_Detection_for_Autonomous_Driving,Deep_MANTA:_A_Coarseto-fine_Many-Task_Network_for_joint_2D_and_3D_vehicle_analysis_from_monocular_image,3D_Bounding_Box_Estimation_Using_Deep_Learning_and_Geometry,M3D-RPN:_Monocular_3D_Region_Proposal_Network_for_Object_Detection,PVNet:_Pixel-Wise_Voting_Network_for_6DoF_Pose_Estimation}. The former uses a neural network to predict the depth map of the monocular image, and converts the depth map into pseudo point clouds by camera calibration files, and then uses a point-cloud-based 3D detection network to regress the 3D boxes. It suffers from increased time and space complexity and decreased detection accuracy owing to the noise produced in the generation of pseudo point clouds. The latter uses prior knowledge to establish the relationship between 2D detection and 3D object, and then regresses the object’s 3D localization in the camera coordinate system. It usually suffers from low accuracy for it has not effectively exploited the relationship between the object’s depth and image scale. 

Aiming at solving the above problems, this paper proposes an end-to-end one-stage 3D detection network based on a Multi-scale Depth-based Stratification structure, MDS-Net. The novelty of the MDS-Net primarily lies in the following three newly obtained features. Firstly, we propose a depth-based stratification structure derived from the conventional Feature Pyramid Network (FPN)~\cite{Feature_Pyramid_Networks_for_Object_Detection}. By establishing mathematical models between the object’s depth and 2D scale, we assign each FPN feature map a different predictable depth range based on prior statistics. Hence, our network inherently has a better depth perception. Secondly, we design a novel angle loss based on the consistency of IoU and angle, which helps the network to predict the more precise boxes and converge faster during training. Thirdly, we design a density-based soft-nms algorithm for the post-processing. Since MDS-Net generates multiple boxes for the same object on different depth stratification, we believe that objects are more likely at locations with a high density of predictions.

In summary, our contributions are as follows:
\begin{itemize}
\item We propose a one-stage monocular 3D object detection network MDS-Net based on a Multi-scale Depth-based Stratification structure, which can accurately predict the object’s localization in the 3D camera coordinate system from the monocular image in an end-to-end manner.
\item We design a novel angle loss function to strengthen the network’s ability of angle prediction.
\item We propose a density-based Soft-NMS method to improve the confidence of more credible boxes.
\item We achieve state-of-art performance on the KITTI benchmark for the monocular image 3D object detection algorithm.
\end{itemize}

\section{Related Work}
\subsection{2D Detection}
At present, with the popularity of convolutional neural networks, a multitude of 2D object detection networks have been raised. Generating proposals by the region proposal network (RPN) has been widely adopted by recent two-stage frameworks. The two-stage approach~\cite{Rich_Feature_Hierarchies_for_Accurate_Object_Detection_and_Semantic_Segmentation,Fast_R-CNN,Faster_R-CNN,R-FCN} first generates the region of interest (ROI) by the RPN and then regresses the object in ROI for refinement. This kind of network is often demonstrated to obtain more accurate detections. The one-stage approach~\cite{Scalable_object_detection_using_deep_neural_networks,You_only_look_once:_Unified_real-time_object_detection,Ssd:_Single_shot_multibox_detector} omits object proposal generation and directly regresses bounding boxes from the image. In this approach, the image is divided into even grids, which will be responsible for predicting the object, if it contains the object’s center point. Compared with the former, one-stage detection is usually considered more efficient. However, due to the lack of RPN, it has a severe imbalance between positive and negative samples. It can also be easily hampered by a wrong-designed hyperparameter of the anchor. To solve this problem, FCOS~\cite{FCOS:_Fully_Convolutional_One-Stage_Object_Detection} and FoveaBox~\cite{FoveaBox:_Beyound_Anchor-Based_Object_Detection} adopted the concept of anchor-free to solve 2D object detection in a per-pixel prediction. They avoid calculating the IoU between GT boxes and anchor boxes, which improves the efficiency and the detection accuracy of the network.

\subsection{3D Detection based on Point Cloud}
Recently, 3D detection based on point clouds~\cite{BirdNet:_A_3D_Object_Detection_Framework_from_LiDAR_Information,BirdNet+,Multi-View_3D_Object_Detection_Network_for_Autonomous_Driving,VoxelNet:_End-to-End_Learning_for_Point_Cloud_Based_3D_Object_Detection,Frustum_PointNets_for_3D_Object_Detection_From_RGB-D_Data,PointNet:_Deep_Learning_on_Point_Sets_for_3D_Classification_and_Segmentation} has developed rapidly. A common theme among SOTA point-based 3D detection methods is to project the point cloud into sets of 2D views. For example, AVOD~\cite{Joint_3D_Proposal_Generation_and_Object_Detection_from_View_Aggregation} projects the 3D point cloud to the bird’s-eye view (BEV), and then fuses the features from BEV and image to predict 3D bounding boxes. Although the view-based method is time-saving, it destroys the information integrity of the point cloud. The voxel-based method~\cite{VoxelNet:_End-to-End_Learning_for_Point_Cloud_Based_3D_Object_Detection} divides the point cloud into regular voxels, and then regresses the 3D bounding boxes by the 3D convolutional network. It preserves the shape information but is usually criticized for its higher time complexity. The set-based method~\cite{Frustum_PointNets_for_3D_Object_Detection_From_RGB-D_Data, PointNet:_Deep_Learning_on_Point_Sets_for_3D_Classification_and_Segmentation} represents the point cloud as a point set and uses a multi-layer perceptron (MLP) to learn features from the unordered set of points directly. Although it reduces the impact of the point cloud’s local irregularity to a certain extent, it still has the problem of mismatch between the regular grid and the point cloud structure.

\subsection{3D Detection based on Monocular}
Monocular 3D object detection still remains a challenge due to the lack of depth information. Most frameworks~\cite{Monocular_3D_Object_Detection_with_Pseudo-LiDAR_Point_Cloud,Multi-Level_Fusion_Based_3D_Object_Detection_From_Monocular_Images,Accurate_Monocular_3D_Object_Detection_via_Color-Embedded_3D_Reconstruction_for_Autonomous_Driving,3DOP} use the imaging geometric relationship to convert the depth map estimated from the image into a pseudo point cloud, and then use the general point-based network to regress the 3D bounding boxes. For example, Xu~\cite{Multi-Level_Fusion_Based_3D_Object_Detection_From_Monocular_Images} used a 2D image network to predict the ROI area, then extracted image features and pseudo point cloud features in the ROI area for 3D prediction. Quite apart from the extra time complexity introduced by depth map prediction, the low accuracy of monocular depth estimation algorithms usually leads to poor prediction effects of such networks. 

More recently, deep neural networks directly processing the RGB image~\cite{Monocular_3D_Object_Detection_for_Autonomous_Driving,Deep_MANTA:_A_Coarseto-fine_Many-Task_Network_for_joint_2D_and_3D_vehicle_analysis_from_monocular_image,3D_Bounding_Box_Estimation_Using_Deep_Learning_and_Geometry,M3D-RPN:_Monocular_3D_Region_Proposal_Network_for_Object_Detection,PVNet:_Pixel-Wise_Voting_Network_for_6DoF_Pose_Estimation} have shown more accurate results. Chen first proposed Mono3D~\cite{Monocular_3D_Object_Detection_for_Autonomous_Driving} to predict the target based on a priori hypothesis of the ground plane. Chabot proposed Deep-MANTA~\cite{Deep_MANTA:_A_Coarseto-fine_Many-Task_Network_for_joint_2D_and_3D_vehicle_analysis_from_monocular_image}, which uses the 3D CAD model to match the object’s key points to estimate 3D dimensions and orientation. M3D-RPN~\cite{M3D-RPN:_Monocular_3D_Region_Proposal_Network_for_Object_Detection} uses prior statistics to initialize the 3D parameter and achieves 3D estimation by a monocular 3D region proposal network. However, none of those mentioned above networks solves the problem that the CNN structure is hard to capture 2D scale changes of the image object caused by depth. 

\begin{figure*}[ht]
	\begin{center}
		\includegraphics[width=0.7\linewidth]{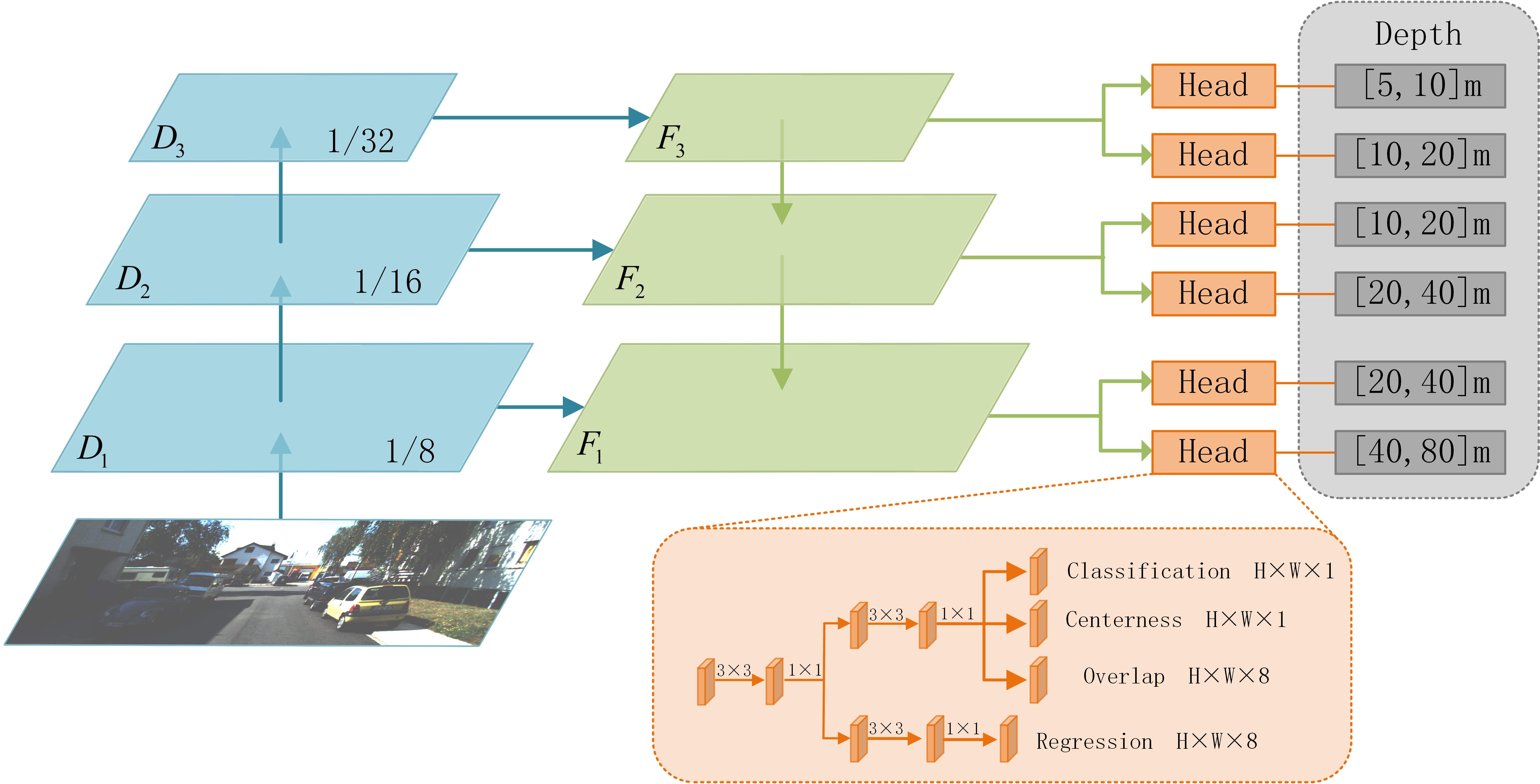}
	\end{center}
	\caption{Overview of our proposed framework.}
	\label{fig:figure1}
\end{figure*}

\section{The Proposed MDS-Net}
This chapter comprises five sections: In section 3.1, we introduce the overall architecture of our network. In section 3.2, we introduce the depth-based stratification structure, which improves monocular depth estimation. In section 3.3, the formulation to transform the network output to 3D coordinate is presented. In section 3.4, we expound on our loss function and propose a novel angle loss to improve angle prediction accuracy. Finally, in section 3.5, we detail a density-based Soft-NMS algorithm to reasonably process the prediction box and increase the recall.

\subsection{Network Architecture}
As shown in Figure \ref{fig:figure1}, Our object detection network consists of three parts: backbone, FPN~\cite{Feature_Pyramid_Networks_for_Object_Detection} and detection head. Our network uses the Darknet53~\cite{You_only_look_once:_Unified_real-time_object_detection} as backbone to generate three feature maps, $D_1, D_2$ and $D_3$, with downsampling ratios of $8, 16$ and $32$, respectively. We then use the FPN to fuse features from different layers and obtain the fused feature maps, $F_1, F_2, F_3$. Finally, each feature map $F_i$ is connected to the detection head. The detection head is composed of two branches. One branch is responsible for predicting the 3D parameters of the bounding box, the other is responsible for predicting the confidence, intersection-over-union (IoU) and center-ness of the object. During the inference, the predicted IoU and center-ness are multiplied with the corresponding confidence to calculate the final score.
\begin{figure}
	\begin{center}
		\includegraphics[width=0.7\linewidth]{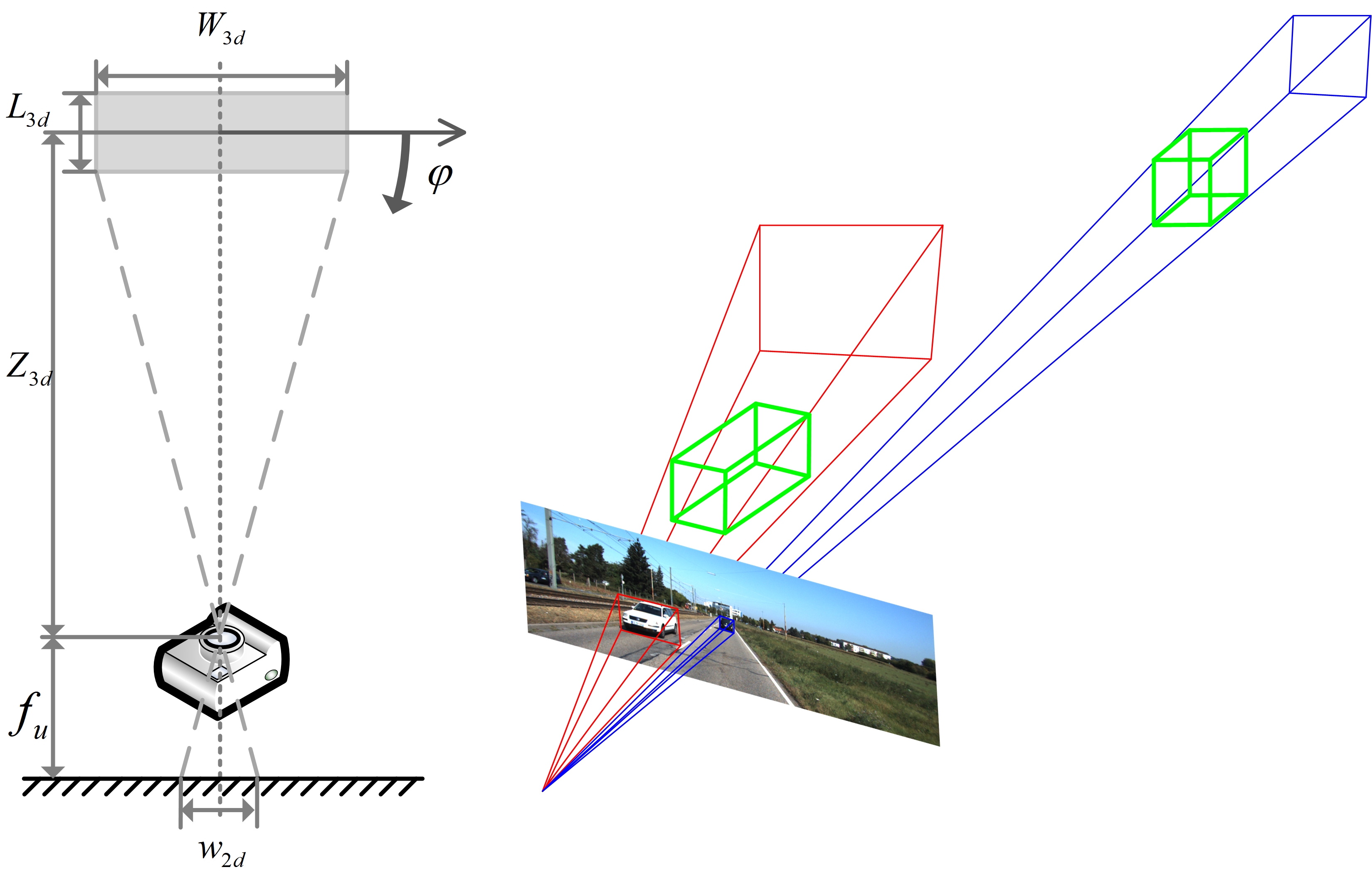}
	\end{center}
	\caption{The relationship between the 3D depth and 2D scale of the object.}
	\label{fig:figure2}
\end{figure}
\subsection{Depth Stratification}
\begin{figure}
	\begin{center}
		\includegraphics[width=0.8\linewidth]{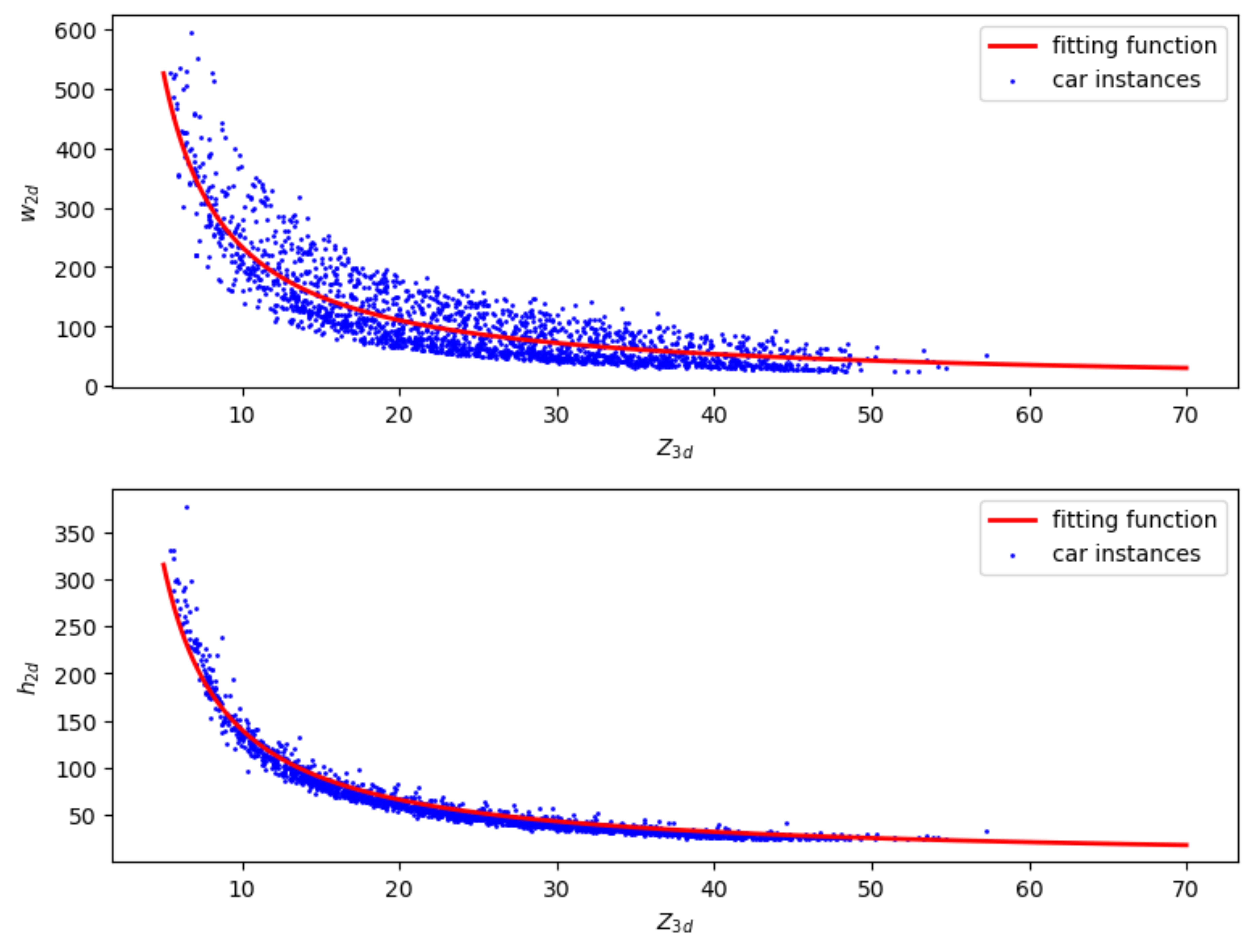}
	\end{center}
	\caption{The statistical relationship between the depth and 2D size of cars on the KITTI train set under hard setting. And we plot fitting function deduced by Equation \ref{eq1}.}
	\label{fig:figure3}
\end{figure}

As shown in Figure \ref{fig:figure2}, we assume that the course angle $\varphi$ is 0, then the relationship between 3D depth $Z_{3d}$ and 2D size $(w_{2d},h_{2d})$ of the object can be approximatively calculated as follows:
\begin{equation}\label{eq1}
	\begin{cases}
		\frac{Z_{3d}-\frac{L_{3d}}{2}}{f_u}=\frac{W_{3d}}{w_{2d}}\\\\
		\frac{Z_{3d}-\frac{L_{3d}}{2}}{f_v}=\frac{H_{3d}}{h_{2d}},
	\end{cases}
\end{equation}
\\
$Z_{3d}$ denotes the object’s depth. $(W_{3d},L_{3d},H_{3d})$ are the 3D sizes of the object. $(w_{2d}, h_{2d})$ are the object’s 2D sizes on the image, and $(f_{u},f_{v})$ are the camera’s focal length. It can be inferred from Equation \ref{eq1} that when focusing on objects of the same kind which having similar 3D sizes like cars, 2D sizes are mainly determined by the depth. But the 2D size of the object is not determined by the depth but is also affected by the course angle. So we then calculated the statistical relationship between the depth and 2D size of cars in KITTI in Figure \ref{fig:figure3} and we can infer that the influence of course angle is minor. Therefore, we establish a mathematical model between depth and 2D size, and apply the FPN structure~\cite{Feature_Pyramid_Networks_for_Object_Detection} to further improve the depth prediction by capturing objects’ size gap on the image. We set $[5, 80]$ meters as the predictable depth range, ignoring the objects outside the range. As shown in Figure \ref{fig:figure1}, we use six detection heads to predict objects in different depth ranges.

Our newly proposed depth-based stratification structure is characterized by the following two properties. Firstly, each feature map’s predictable depth range is increased by a geometric growth rate. For instance, $F_2$ can predict two times the depth range of $F_1$, since the receptive field of $F_2$ is approximately twice bigger than $F_1$. It is wise to keep the growth rate of the feature map’s predictable depth range consistent with the receptive fields for a better network depth prediction. Secondly, three feature maps’ predictable depth ranges actually overlap each other. This is caused namely by the two reasons. On the one hand, from Equation \ref{eq1}, we can deduce that the 2D size of the object is not strictly inversely proportional to the depth but is also affected by the object’s pose. We need to expand each feature map’s predictable depth range to strengthen network prediction’s robustness. On the other hand, our overlapping stratification structure enables the network to predict targets in crucial areas multiple times, which in consequence eases the imbalance between positive and negative samples to a certain extent. So we use $F_1$, $F_2$ and $F_3$ to predict objects in the range of $5-20m$, $10-40m$ and $20-80m$ respectively.

\subsection{Network Predictions}
The outputs of our network consist of two parts: the predicted 3D parameters of an object, including its 3D center, 3D size and angle, the predicted score of an object, including its confidence, IoU and center-ness.
\\
\textbf{3D Prediction:} According to the pinhole model of the camera, we project the object’s 3D center $[X_{3d},Y_{3d},Z_{3d}]^T$ in the camera coordinates into the image.
\begin{equation}\label{eq2}
	\begin{vmatrix}
		f_u& 0  & c_u\\
		0  & f_v& c_v\\
		0  &  0& 1 
	\end{vmatrix}
	\times
	\begin{vmatrix}
		X_{3d}\\
		Y_{3d}\\
		Z_{3d} 
	\end{vmatrix}
	=Z_{3d}\begin{vmatrix}
		u\\
		v\\
		1 
	\end{vmatrix},
\end{equation}
\\
where $(u,v)$ denote the coordinates of the projected center in the pixel coordinates. $(f_u,f_v)$ are the focal length and $(c_u,c_v)$ are the principal point offset of the camera.

According to the Equation \ref{eq2}, we can obtain the relationship between $(u,v)$ and $(X_{3d},Y_{3d},Z_{3d})$ as follows:
\begin{equation}\label{eq3}
	\begin{cases}
		X_{3d}=\frac{(u-c_u)Z_{3d}}{f_u}\\\\
		Y_{3d}=\frac{(v-c_v)Z_{3d}}{f_v},
	\end{cases}
\end{equation}
\\
where $c_u,c_v,f_u,f_v$ are the parameters of the camera, $u,v,Z_{3d}$ can be predicted by the network.
\begin{figure}
	\centering
	\subfigure[]{
		\label{fig:fig4a}
		\includegraphics[width=0.45\linewidth]{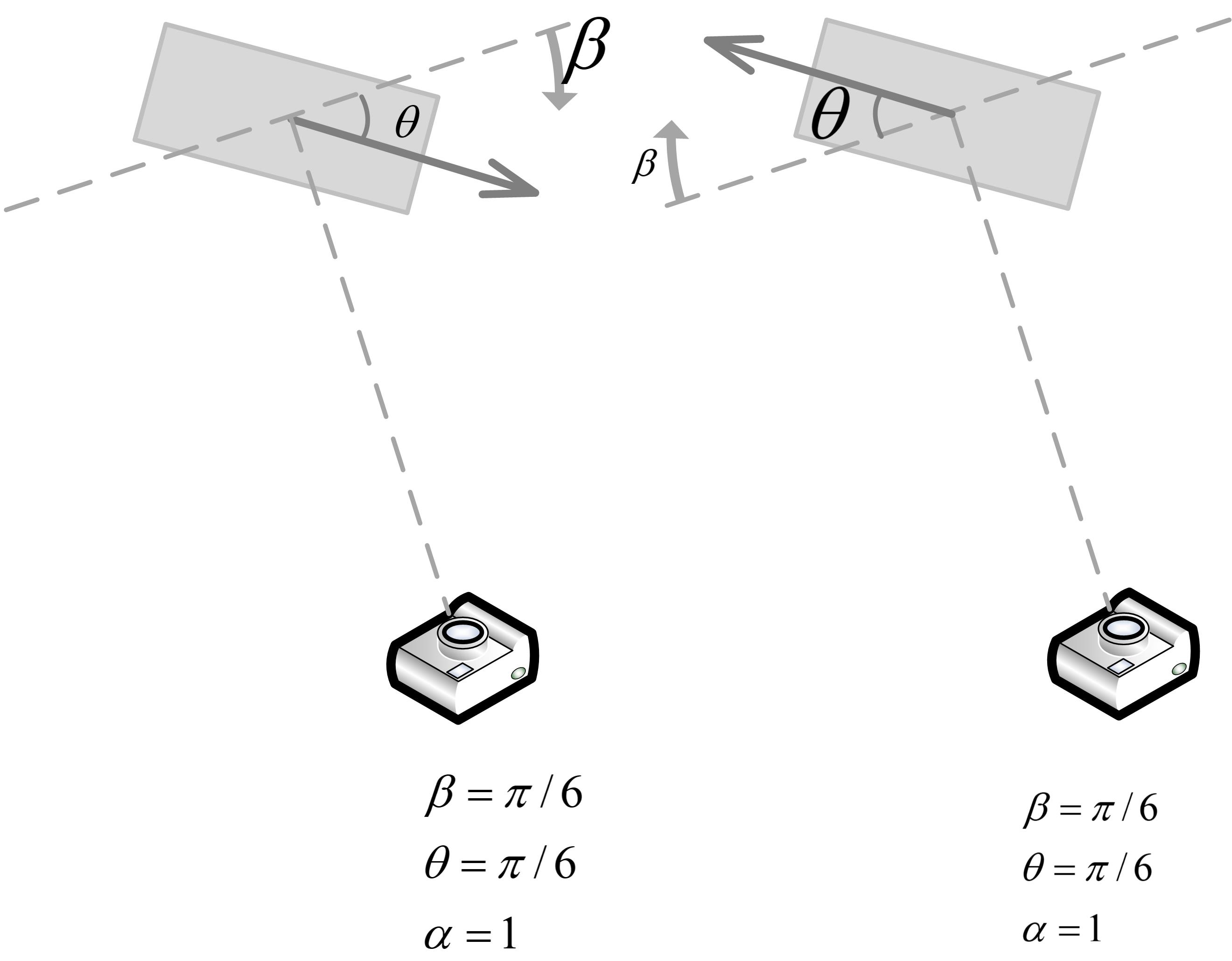}
	}
	\subfigure[]{
		\label{fig:fig4b}
		\includegraphics[width=0.45\linewidth]{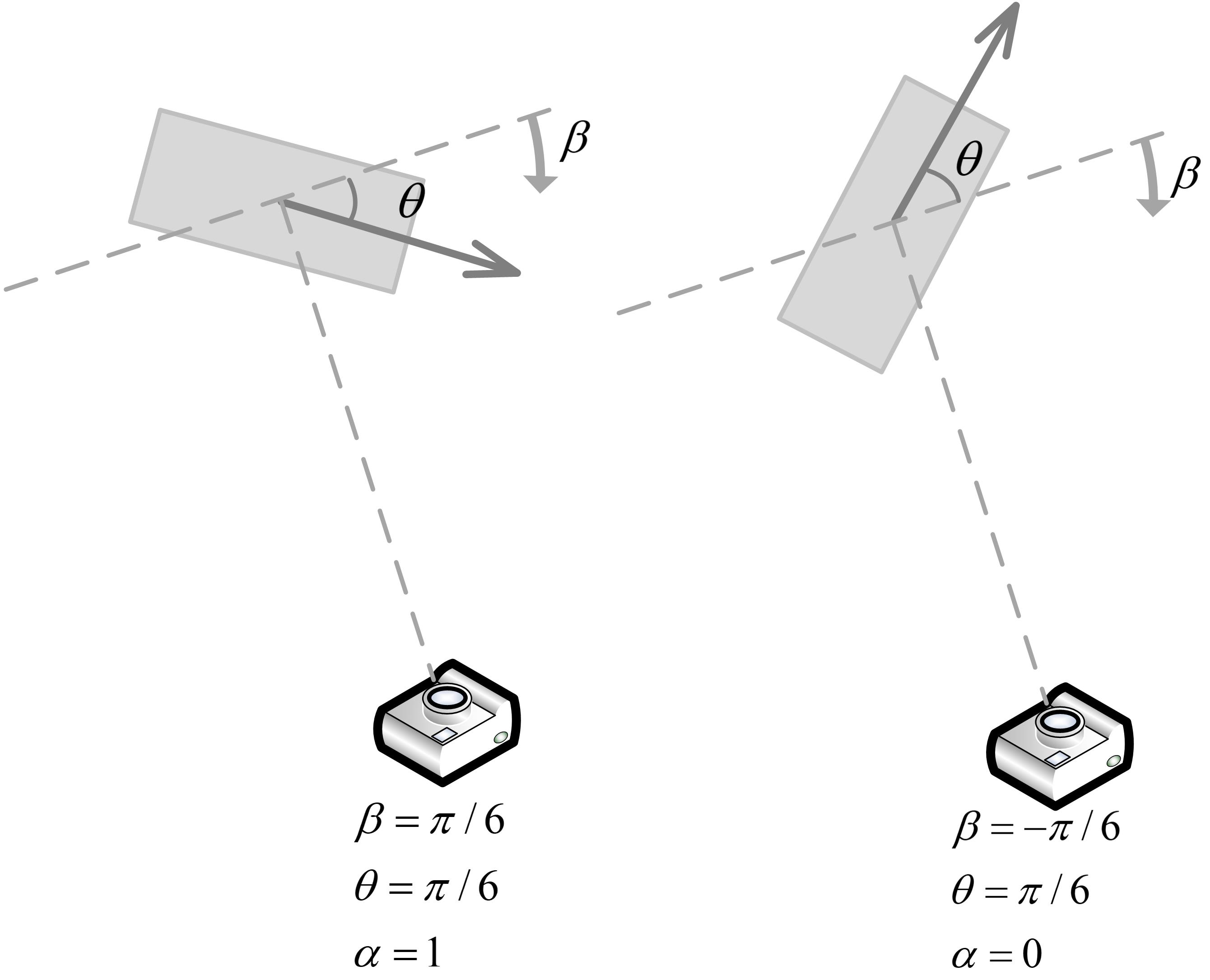}
	}
	\caption{The observation angle specifications in our network: (a) shows the two states of the 3D box correspond to the same angle value. (b) shows the operation of splitting $\beta$ into object heading $\alpha$ and angle offset $\theta$.}
\end{figure}
For the 3D size prediction, we calculate the average 3D size $(w_0, l_0, h_0)$ of all objects in the data set as the preset value and then predict the logarithm of the ratio between the ground truth and the preset value. We use the observation angle, which is intuitively more meaningful when processing image features, in the angle prediction. The range of the observation angle $\beta$ is $(-\pi/2,\pi/2)$. As shown in Figure \ref{fig:fig4a}, the two states of the 3D boxes correspond to the same angle value $\beta$ in our algorithm. As shown in Figure \ref{fig:fig4b}, we split $\beta$ into the object heading $\alpha$ (forward or backward) and angle offset $\theta$ for prediction. 

Suppose the predicted values on the feature map $F_i$ are $\hat{u},\hat{v},\hat{Z},\hat{w},\hat{l},\hat{h},\hat{\alpha},\hat{\theta}$, then we use the following formula to calculate the 3D parameters $X_{3d},Y_{3d},Z_{3d},W_{3d},L_{3d},H_{3d},\beta$ of the object:
\begin{equation}\label{eq4}
	\begin{aligned}	
		&\begin{cases}
			Z_{3d}&=Z_i\cdot{2^{\hat{Z}}},\\
			X_{3d}&=\frac{[X_P+{\hat{u}}-c_u](Z_i\cdot{2^{\hat{Z}}})}{f_u},\\
			Y_{3d}&=\frac{[Y_P+{\hat{v}}-c_v](Z_i\cdot{2^{\hat{Z}}})}{f_v},\\
		\end{cases}\\	
		&\begin{cases}
			W_{3d}&=e^{\hat{w}}\cdot w_0,\\
			L_{3d}&=e^{\hat{l}}\cdot l_0,\\
			H_{3d}&=e^{\hat{h}}\cdot h_0,\\
		\end{cases}\quad
		\beta = \begin{cases}
			\hat{\theta}, if \quad \hat{\alpha}=1 \\
			-\hat{\theta}, if \quad \hat{\alpha}=0, \\
		\end{cases}&
	\end{aligned}
\end{equation}
\\
where $Z_i$ is the minimum depth that the detection head needs to regress, $(X_P,Y_P)$ represent the 2d center of the grid.
\\
\textbf{Score Prediction:} 
For the confidence prediction, according to KITTI’s standard~\cite{Are_we_ready_for_autonomous_driving?_The_KITTI_vision_benchmark_suite} for difficulty level, we piecewise predict the confidence of different difficulty levels and set the ground truth to $1.0, 0.8, 0.6, 0.4$, respectively. Following FCOS~\cite{FCOS:_Fully_Convolutional_One-Stage_Object_Detection}, the center-ness is defined as the normalized distance from the grid’s center to the object’s 2D center. Following IoUNet~\cite{Acquisition_of_Localization_Confidence_for_Accurate_Object_Detection}, the IoU is defined as the intersection over the union of the prediction box and the ground-truth box on the bird’s eye view.

\subsection{Loss}
We set the pixels inside the projected ground-truth box as positive samples and those outside as negative. Considering that our network uses the depth-based stratification structure, we set confidence label of a pixel as “ignore” if it lies within an object beyond the predictable depth range of the relative feature map. As the solution to the label ambiguity that occurs when two or more boxes overlap on the image, we set pixel only responsible for the closest object due to the visibility in the camera’s perspective.

The loss $L$ in our network is composed of classification loss $L_c$ and 3D box loss $L_{3d}$: 

\begin{equation}\label{eq5}
	L=L_c+L_{3d}.
\end{equation}
We use the sum of confidence loss, center-ness loss and IoU loss as the classification loss, and ignore the contribution from all samples labeled as “ignore”. Inspired by FCOS~\cite{FCOS:_Fully_Convolutional_One-Stage_Object_Detection}, we add two branches parallel with the confidence branch to predict the center-ness and IoU, and use the Quality Focal Loss (QFL)~\cite{Generalized_Focal_Loss:_Learning_Qualified_and_Distributed_Bounding_Boxes_for_Dense_Object_Detection} to optimize the confidence, center-ness and predicted IoU.

\begin{figure}
	\centering
	\subfigure[]{
		\label{fig:fig5a}
		\includegraphics{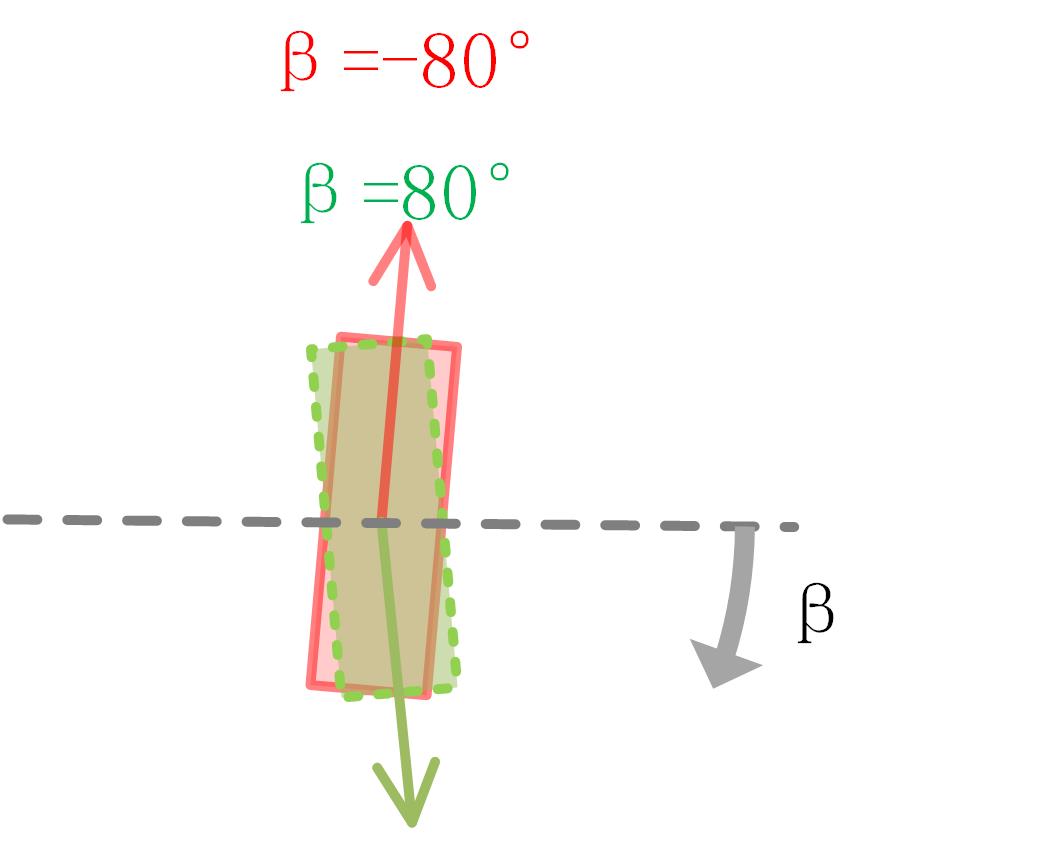}
	}
	\subfigure[]{
		\label{fig:fig5b}
		\includegraphics{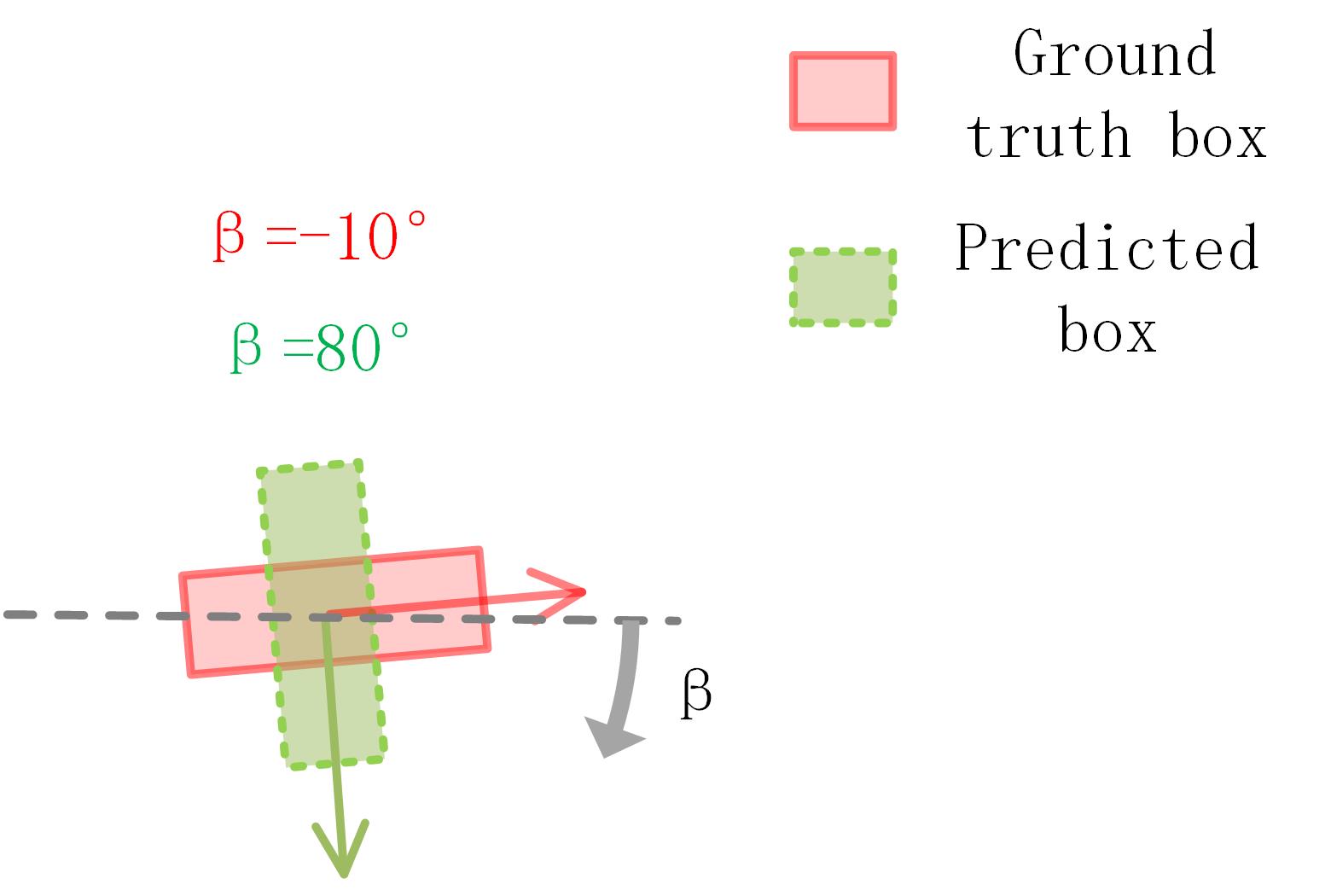}
	}
	\caption{ Two cases where the angle loss function used in previous work may encounter problems.}
\end{figure}

We add up localization loss, size loss and angle loss as our $L_{3D}$ loss. We use the L2 loss to optimize the localization and size predictions. VoxelNet~\cite{VoxelNet:_End-to-End_Learning_for_Point_Cloud_Based_3D_Object_Detection} directly uses the offset of radians as the loss function. However, in the case shown in Figure \ref{fig:fig5a}, the overlap of the two boxes is considerable while the network still generates a large angle loss. To solve this problem, SECOND~\cite{SECOND:_Sparsely_Embedded_Convolutional_Detection} proposed a new angle loss as follows:
\begin{equation}\label{eq6}
	L_{rot}=SmoothL1[sin(\beta-\hat{\beta})],
\end{equation}
\\
where $\beta,\hat{\beta}$ denote ground truth observation angle and the predicted observation angle, respectively. Although this function naturally models IoU against the angle offset function, in the case shown in Figure \ref{fig:fig5b}, the derivative of the loss is improperly small despite the loss reaching the maximum value, making it difficult for the network to regress. To overcome the above problems, we design a new angle loss:

\begin{equation}\label{eq7}
	L_{rot}=(\theta-\hat{\theta})^2+(\alpha-\hat{\alpha})^2\cdot sin(2\theta),
\end{equation}
\\
where $\alpha,\hat{\alpha}$ denote ground truth heading and the predicted heading, $\theta,\hat{\theta}$ denote ground truth angle offset and the predicted angle offset, respectively. Our angle loss not only establishes a consistent model between the IoU and the observation angle but also makes it easier to train the network when the angle loss rises to the maximum. 
\begin{figure}
	\begin{center}
		\includegraphics[width=1\linewidth]{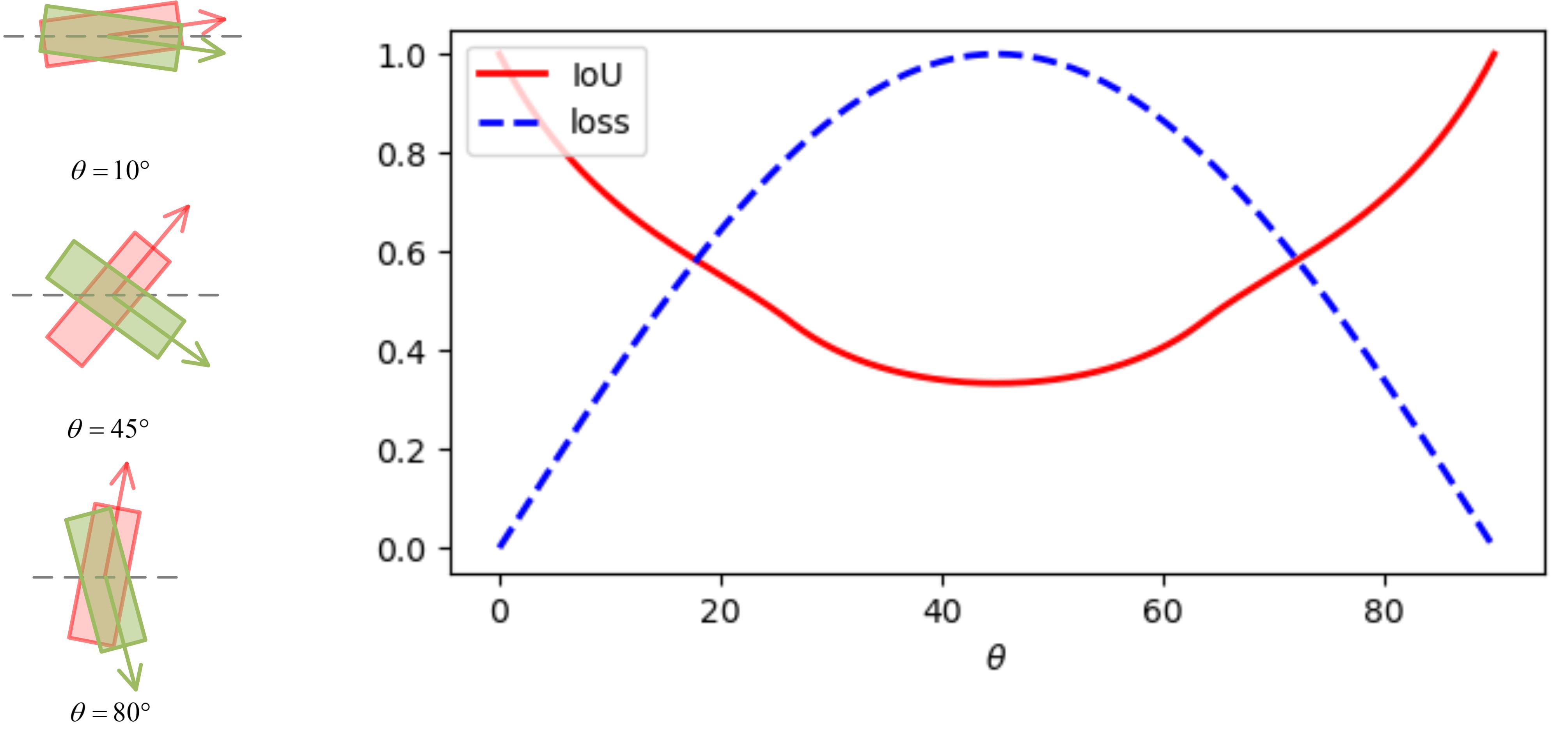}
	\end{center}
	\caption{Left: The influence of heading prediction error on IoU at different values of $\theta$. Right: The relationships among loss, IoU with $\theta$.}
	\label{fig:figure7}
\end{figure}

We assume that the network has predicted an accurate $\hat{\theta}$. As shown in Figure \ref{fig:figure7}, the inaccurately predicted heading has a minor effect on the IoU when the ground truth angle offset is close to either $0^{\circ}$ or $90^{\circ}$, while it has a significant impact on the IoU when the ground truth angle is close to $45^{\circ}$. We apply a weighted parameter $sin(2\theta)$ to the heading loss to increase the penalty for the inaccurate heading prediction when the ground truth angle is close to $45^{\circ}$.
\begin{table*}[t]
	\setlength{\tabcolsep}{4.0mm}
	\centering
	\begin{tabular}{c|c|ccc|ccc}\hline
		& \multirow{2}*{\makecell[c]{Time\\(sec)}} & \multicolumn{3}{c|}{$AP_{BEV}$} & \multicolumn{3}{c}{$AP_{3D}$}\\
		&  & Easy & Mod & Hard & Easy & Mod & Hard\\\hline
		Shift R-CNN~\cite{shift_r-cnn}& 0.25&11.84&6.82&5.27&6.88&3.87&2.83\\
		MonoGRNet~\cite{monogrnet}& 0.04&18.19&11.17&8.73&9.61&5.74&4.25\\
		M3D-RPN~\cite{M3D-RPN:_Monocular_3D_Region_Proposal_Network_for_Object_Detection}&	0.16&21.02&	13.67&10.23&14.76&9.71&7.42\\
		RTM3D~\cite{li2020rtm3d} &0.05&19.17 & 14.20& 11.99 &14.41 & 10.34 & 8.77 \\
		MonoPair~\cite{chen2020monopair}&0.06& 19.28 & 14.83& 12.89 &13.04& 9.99 & 8.65\\
		MonoRCNN~\cite{shi2021geometry} & 0.07 &25.48 & \textbf{\textit{18.11}}& \textbf{\textit{14.10}}&18.36 & 12.65 & \textbf{\textit{10.03}} \\
		DA-3Ddet ~\cite{ye2020monocular} &0.4& 23.35 & 15.90& 12.11& 16.77 & 11.50 & 8.93\\
		D$^4$LCN ~\cite{ding2020learning} & 0.2&22.51& 16.02& 12.55& 16.65 & 11.72 & 9.51 \\
		Kinematic3D ~\cite{brazil2020kinematic} &0.12& 26.69 & 17.52 &13.10& 19.07 & 12.72& 9.17\\
		Ground-Aware~\cite{liu2021ground}&0.05&\textbf{\textit{29.81}} &	17.98 &13.08& \textbf{\textit{21.65}} &	\textbf{\textit{13.25}} &	9.91\\
		YOLOMono3D ~\cite{liu2021yolostereo3d} &0.05&26.79&17.15&12.56&	18.28&	12.06 &8.42 \\\hline\hline
		Ours&0.05&\textbf{30.92}&\textbf{18.65}&\textbf{14.53}&\textbf{22.8}&\textbf{13.4}&\textbf{10.27}\\\hline
		
	\end{tabular}

	\caption{Comparison of our framework to image-only 3D localization frameworks on the Bird’s Eye View task and 3D object detection task. (The time is reported from the official leaderboard with slight variances in hardware).We use \textbf{bold type} to indicate the highest result and \textbf{\textit{italic type}} for the second-highest result.}\label{table1}
\end{table*}
\begin{table}[t]
	\setlength{\tabcolsep}{1.5mm}
	\centering
	\begin{tabular}{c|ccc}\hline
		&  \multicolumn{3}{c}{IoU$\geq$0.7 / IoU$\geq$0.5}\\
		&   $5-20m$ & $10-40m$ & $20-80m$ \\\hline
		M3D-RPN	&20.63/60.63 &	11.43/38.79&	3.36/19.14\\
		Ours& 30.61/73.22 	&18.74/54.92	&6.31/30.07 \\\hline\hline
		\textit{Improvement}&+9.98/+12.59 	&+7.31/+16.13	&+2.95/+10.93\\\hline	
	\end{tabular}
	\caption{The 3D detection results for Car in different depth ranges on KITTI  validation  set~\cite{VAL1}.}\label{table2}
\end{table}
\begin{table}[t]
	\setlength{\tabcolsep}{1.8mm}
	\centering
	\begin{tabular}{c|ccc|ccc}\hline
		& \multicolumn{3}{c|}{Pedestrian[IoU$\geq$0.5]} & \multicolumn{3}{c}{Cyclist[IoU$\geq$0.5]}\\
		&  Easy & Mod & Hard & Easy & Mod & Hard\\\hline

		M3D-RPN&4.75&	3.55&	2.79&	3.10&	1.49&	1.17\\	
		Ours&9.04&	6.27&	4.91&	3.17&	1.80&	1.64\\\hline
	\end{tabular}
	\caption{The 3D detection results of MDS-Net for Pedestrian and Cyclist on KITTI validation set.}\label{table5}
\end{table}

\subsection{Density-based Soft-NMS}
Our network will predict multiple boxes for one object in different depth stratifications. The traditional Soft-NMS~\cite{Soft-NMS} algorithm selects the detection box $M$ according to the predicted confidence score in descending order. It decays the confidence score of box $b_i$ that has a high overlap with $M$, as follows:
\begin{equation}\label{eq8}
	\tilde{s}_i=s_ie^{-\frac{iou(M,b_i)^2}{\sigma}},
\end{equation}
\\
where $s_i$ denotes the original score of $b_i$, $\tilde{s}_i$ denotes the updated score after implementing the Soft-NMS algorithm, $\sigma$ is a hyperparameter of decay coefficient.

Following the idea, we develop a density-based Soft-NMS algorithm to filter repeated boxes. The key strategy of our NMS algorithm is that the denser the predicted box is, the more likely the object exists. We define the density of a candidate box as the sum of IOU between the candidate box and all surrounding boxes. Based on the results obtained by the traditional Soft-NMS algorithm~\cite{Soft-NMS}, we activate the confidence of the candidate box with a high density:
\begin{equation}\label{eq9}
	\tilde{s}_m=s_m(2-e^{-\frac{\sum_{b_i\in B}iou(M,b_i)^2}{\gamma}}),
\end{equation}
\\
where $B$ is the set of all boxes, $s_m$ is the original score of the box $M$, $\tilde{s}_m$ is the renewed score after implementing our NMS, and $\gamma$ is another hyperparameter of decay coefficient.
\section{Experimental Results}
\subsection{Dataset}
We evaluate our MDS-Net on the challenging KITTI data set~\cite{Are_we_ready_for_autonomous_driving?_The_KITTI_vision_benchmark_suite}. The 3D detection of this data set contains two core tasks: bird’s eye view (BEV) object detection and 3D object detection. The dataset assigns its samples to three difficulty categories (easy, medium and difficult) according to the object’s truncation, occlusion and 2D bounding box height. We evaluate our algorithm on KITTI’s three classes in this work. Following ~\cite{Accurate_Monocular_3D_Object_Detection_via_Color-Embedded_3D_Reconstruction_for_Autonomous_Driving}, we apply the two split methods of validation~\cite{VAL1} and official test splits~\cite{Are_we_ready_for_autonomous_driving?_The_KITTI_vision_benchmark_suite}, In each split method, the data from the same sequence will only appear in one split so that the interference of adjacent frames to the network model is eliminated.

Following ~\cite{Disentangling_Monocular_3D_Object_Detection}, KITTI uses Average Precision on $40$ recall positions $(AP_{40})$ on the official test splits to replace Average Precision on $11$ recall positions $(AP_{11})$. On the official test split, we show the $AP_{40}$ score with a $0.7$ IoU threshold. 
  
\subsection{Implement Detail}
Seeing that the center-ness and IoU prediction need a better box prediction result, we first train the location regression network except for the center-ness and IoU branches with a piecewise decayed learning rate. The initial learning rate is $10^{-4}$, and it will decay to $3\cdot 10^{-5}$ and $10^{-5}$ at the $30^{th}$ epoch and $50^{th}$ epoch, respectively. We then train the score branch for 10 epochs with a learning rate of $10^{-4}$. We use the pre-trained weights under the COCO dataset~\cite{COCO} to initialize our backbone. In our density-based Soft-NMS algorithm we set $\sigma = 0.9$ and $\gamma = 20$ for post-processing.
\begin{figure}[t]
	\begin{center}
		\includegraphics[width=0.85\linewidth]{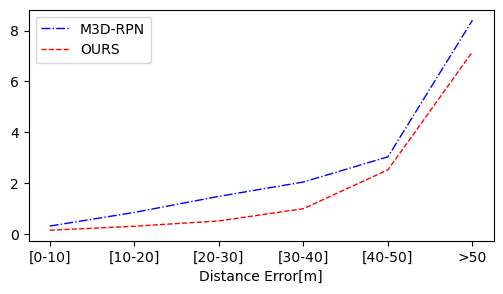}
	\end{center}
	\caption{Average depth estimation error visualized in intervals of $10 m$}
	\label{fig:figure8}
\end{figure}
\begin{figure*}[t]
	\begin{center}
		\includegraphics[width=1.0\linewidth,height=0.22\textheight]{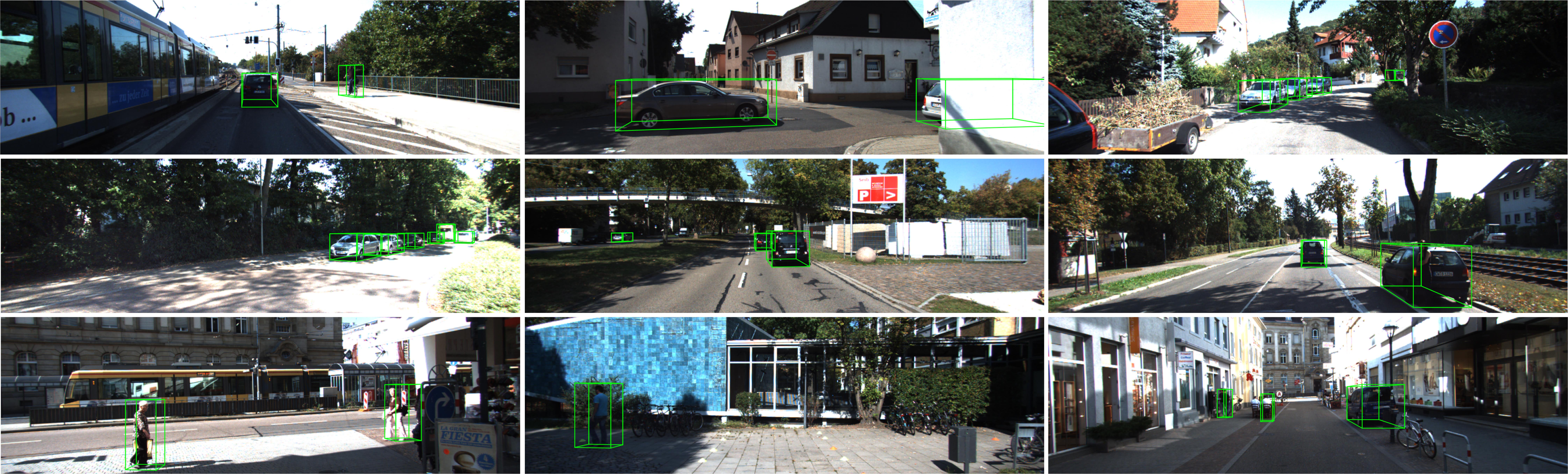}
	\end{center}
	\caption{Qualitative results of our framework in the KITTI validation set.}
	\label{fig:figure9}
\end{figure*}

\begin{table*}[t]
	\centering
	\setlength{\tabcolsep}{4.2mm}
	\begin{tabular}{ccc|ccc|ccc}\hline
		\multirow{2}*{\makecell[c]{depth\\stratification}}&\multirow{2}*{\makecell[c]{density-based\\ Soft-NMS}}&\multirow{2}*{\makecell[c]{piecewise\\ confidence}}&\multicolumn{3}{c|}{$AP_{3D}$}&	\multicolumn{3}{c}{$AP_{BEV}$}\\		
		& & &Easy & Mod & Hard & Easy & Mod & Hard\\\hline
		& & &0.24&	0.41&	0.31&	1.20&	2.08&	1.84\\
		\checkmark& & &11.13&	7.30&	6.13&	16.99&	11.49&	8.92\\
		\checkmark&\checkmark& &12.41&	11.35&	9.86&	18.11&	16.56&	14.81\\
		\checkmark& &\checkmark&17.52&	9.28&	6.82&	24.91&	13.85&	10.29\\
		\checkmark&\checkmark&\checkmark&24.09&	15.49&	12.24&	32.74&	21.25&	17.06\\\hline
	\end{tabular}
	\caption{Ablation Experiments on network components.}\label{table3}
\end{table*}
\begin{table}[t]
	\centering
	\setlength{\tabcolsep}{1mm}
	\centering
	\begin{tabular}{c|ccc|ccc}\hline
		&	\multicolumn{3}{c|}{$AP_{3D}$}&	 \multicolumn{3}{c}{$AP_{BEV}$}\\
		&Easy & Mod & Hard & Easy & Mod & Hard\\\hline
		AVOD&	3.13&	2.06&	1.66&	5.12&	3.57&	2.96\\
		SECOND&	22.19&	14.31&	11.29&	31.02&	20.03&	16.14\\
		OURS&	24.09&	15.49&	12.24&	32.74&	21.25&	17.06\\\hline
	\end{tabular}
	\caption{Comparison of our angle loss function with AVOD~\cite{Joint_3D_Proposal_Generation_and_Object_Detection_from_View_Aggregation} and SECOND~\cite{SECOND:_Sparsely_Embedded_Convolutional_Detection} on the KITTI validation set.}\label{table4}
\end{table}
\subsection{Results}
We evaluate our network for the BEV object detection task and the 3D object detection task, each under both validation~\cite{VAL1} and the official test dataset~\cite{Are_we_ready_for_autonomous_driving?_The_KITTI_vision_benchmark_suite}. In Table \ref{table1}, we compare the evaluation results of our network with the existing state-of-art monocular 3D detection algorithms. Our network obtains remarkable results on the Car detection of all three difficulty levels, especially the easy level, for both tasks. For instance, under test data split with $IoU \geq 0.7$, we surpass the previous state-of-the-art BEV object detection approach by $+1.11\%$ on easy, $+0.54\%$ on moderate and $+0.43\%$ on hard and surpass the previous state-of-the-art 3D object detection approach by $+1.15\%$ on easy, $+0.15\%$ on moderate and $+0.24\%$ on hard.

And then we evaluate on car class in different depth ranges $(5-20m, 10-40m, 20-80m)$ with an IoU criteria of $0.7$ and $0.5$ on the KITTI validation set to prove the validity of depth stratification. In Table \ref{table2}, our method outperforms M3D-RPN\cite{M3D-RPN:_Monocular_3D_Region_Proposal_Network_for_Object_Detection} with margins on AP of $+9.98\%/+12.59\%$ in $5-20m$, $+7.31\%/+16.13\%$ in $10-40m$ and $+2.95\%/+10.93\%$ in $20-80m$ for criteria of $IoU \geq 0.7$ and $IoU \geq 0.5$. We find the model achieves relatively low results in $20-80m$ for criteria of $IoU \geq 0.7$. The results are counterintuitive because the main purpose of FPN features is to help detect small and hard objects in general. So we display average depth estimation error in intervals of $10$ meters in Figure \ref{fig:figure8}. We calculate the difference of the depth between every ground truth object and its best-predicted box. As shown in Figure \ref{fig:figure8}, the depth estimation error increases as the distance grows. Therefore, although we achieve a more accurate depth estimation in $50-80m$, we can't achieve high performance in $50-80m$ like $0-50m$ under severe criteria of $IoU \geq 0.7$. This also raises the question of whether the KITTI’s evaluation metric is reasonable for far objects in the monocular 3D object detection task.

And also we compare the results of our method with M3D-RPN\cite{M3D-RPN:_Monocular_3D_Region_Proposal_Network_for_Object_Detection} for pedestrian and cyclist class on the KITTI validation set in Table \ref{table5}. It must be noted that by regressing the 3D bounding boxes on the image directly, our approach reaches the operating speed of $20$ FPS and better meets the real-time requirements of autonomous driving, as compared to the existing networks based on the pseudo point cloud. 

In Figure \ref{fig:figure9}, we visualize several 3D detection results on the KITTI validation set~\cite{VAL1}.

\subsection{Ablation Study}
For the ablation study, following ~\cite{VAL1}, we divide the training samples of the KITTI data set into $3712$ training samples and $3769$ verification samples and then verify the network’s accuracy according to the $AP_{40}$ standard.
\\
\textbf{Depth Stratification:} We compare the object detection $AP_{40}$ with and without depth stratification. In the test without depth stratification, by referring to the strategy of the yolov3 network~\cite{You_only_look_once:_Unified_real-time_object_detection}, we use IoU between the 2D box of the object and the preset anchor as the basis for stratification. As shown in the first and second rows of Table \ref{table3}, the network based on Multi-scale Depth Stratification (MDS) (the second row) achieves a significant gain of $6.89\%$ over the baseline implementation (the first row) on 3D detection, which verifies the superiority of the MDS structure.
\\
\textbf{Density-based Soft-NMS and Piecewise Score:} To better understand the effect of the density-based soft-NMS and piecewise score, we ablate them by using a standard soft-NMS~\cite{Soft-NMS} and setting all object scores to $1$ as the baseline implementation (the second row in Table \ref{table3}). We observe that both components achieve a considerable gain in both 3D and BEV perspectives. The combination of the two components surpasses the baseline by $8.19\%$ in the 3D detection.
\\
\textbf{Angle Loss:} We compare the 3D and BEV detection AP of our proposed angle loss with the angle loss of AVOD~\cite{Joint_3D_Proposal_Generation_and_Object_Detection_from_View_Aggregation} and that of SECOND~\cite{SECOND:_Sparsely_Embedded_Convolutional_Detection}. It shows that our angle loss focuses more on the consistency of angle prediction and boxes overlap, rather than struggling with numerical prediction. As shown in Table \ref{table4}, our angle loss improves performance by $1.22\%$ in and $1.18\%$ in over the SECOND~\cite{SECOND:_Sparsely_Embedded_Convolutional_Detection}. We also observe from the training process in practice that our angle loss improves the convergence speed of the angle regression, suggesting that we mitigate the problem of extrema convergence in the SECOND~\cite{SECOND:_Sparsely_Embedded_Convolutional_Detection}.

\section{Conclusion}
In this work, we propose a one-stage anchor-free network, which realizes direct detection of 3D objects from monocular images by applying the newly proposed depth-based stratification structure, iou-aware angle loss and density-based Soft-NMS algorithm. The experimental results based on the KITTI dataset~\cite{Are_we_ready_for_autonomous_driving?_The_KITTI_vision_benchmark_suite} have shown that our network can significantly improve the accuracy and recall rate in the bird’s-eye view object detection tasks and 3D object detection tasks.

{
	\bibliography{egbib}
}
\end{document}